\documentclass{article}

\usepackage{times}
\usepackage{graphicx} 
\usepackage{subfigure}

\usepackage{url}

\usepackage[turnoff]{notes} 
\usepackage{natbib}
\usepackage{amsmath}
\usepackage{amssymb}
\usepackage{amsthm}
\usepackage{algorithm}
\usepackage{algorithmic}

\usepackage{hyperref}

\usepackage[accepted]{icml2014} 

\newcommand{\vect}[1]{\boldsymbol{#1}}
\DeclareMathOperator*{\argmax}{arg\,max}

\icmltitlerunning{Two-Stage Metric Learning}

\begin{document} 

\twocolumn[
\icmltitle{Two-Stage Metric Learning}

\icmlauthor{Jun Wang}{jun.wang@unige.ch}
\icmladdress{Department of Computer Science, University of Geneva, Switzerland}
\icmlauthor{Ke Sun}{ke.sun@unige.ch}
\icmladdress{Department of Computer Science, University of Geneva, Switzerland}
\icmlauthor{Fei Sha}{feisha@usc.edu}
\icmladdress{Department of Computer Science, University of Southern California, Los Angeles, CA, USA}
\icmlauthor{Stephane Marchand-Maillet}{Stephane.Marchand-Maillet@unige.ch}
\icmladdress{Department of Computer Science, University of Geneva, Switzerland}
\icmlauthor{Alexandros Kalousis}{Alexandros.Kalousis@hesge.ch}
\icmladdress{Department of Business Informatics,University of Applied Sciences,Western Switzerland,\\
Department of Computer Science, University of Geneva, Switzerland}
\icmlkeywords{boring formatting information, machine learning, ICML}

\vskip 0.3in
]

\begin{abstract} 
In this paper, we present a novel two-stage metric learning algorithm. We first map each learning instance to a probability distribution by computing its similarities to a set of fixed anchor points. Then, we define the distance in the input data space as the Fisher information distance on the associated statistical manifold. This induces in the input data space a new family of distance metric with unique properties. Unlike kernelized metric learning, we do not require the similarity measure to be positive semi-definite. Moreover, it can also be interpreted as a local metric learning algorithm with well defined distance approximation. We evaluate its performance on a number of datasets. It outperforms significantly other metric learning methods and SVM. 
\end{abstract} 

\section{Introduction}
\label{sec:int}

Distance measures play a crucial role in many machine learning tasks and algorithms. Standard distance metrics, e.g. Euclidean, cannot address in a satisfactory manner the multitude of learning problems, a fact that led to the development of metric learning methods which learn problem-specific distance measure directly from the data~\cite{weinberger2009distance,junwang2012,jain2010inductive}. Over the last years various metric learning algorithms have been shown to perform well in different learning problems, however, each comes with its own set of limitations.



Learning the distance metric with one global linear transformation is called single metric learning~\cite{weinberger2009distance,davis2007information}. In this approach the distance computation is equivalent to applying on the learning instances a learned linear transformation followed by a standard distance metric computation in the projected space. 
Since the discriminatory power of the input features might vary locally, this approach is often not flexible enough to fit well the distance in different regions.



Local metric learning addresses this limitation by learning in each neighborhood one local metric~\cite{nohgenerative,junwang2012}. 
When the local metrics vary smoothly in the feature space, learning local metrics is equivalent to learning the Riemannian metric on the data manifold~\cite{Soren2012}.
The main challenge here is that the geodesic distance endowed by the Riemannian metric is often computationally very expensive. In practice, it is approximated by assuming that the geodesic curves are formed by straight lines and the local metric does not change along these lines~\cite{nohgenerative,junwang2012}. Unfortunately, the approximation does not satisfy the symmetric property and therefore the result is a non-metric distance. 

Kernelized Metric Learning (KML) achieves flexibility in a different way ~\cite{jain2010inductive,wang2011metric}. In KML learning instances are first mapped into the Reproducing-Kernel Hilbert Space (RKHS) by a kernel function and then a global Mahalanobis metric is learned in the RKHS space. By defining the distance in the input feature space as the Mahalanobis distance in the RKHS space, KML is equivalent to learning a flexible non-linear distance in the input space. 
However, its main limitation is that the kernel matrix induced by the kernel function must be Positive Semi-Definite (PSD). Although Non-PSD kernel could be transformed into PSD kernel~\cite{chen2008training,ying2009analysis}, the new PSD kernel nevertheless cannot keep all original similarity information.

In this paper, we propose a novel two-stage metric learning algorithm, Similarity-Based Fisher Information Metric Learning (SBFIML). 
It first maps instances from the data manifold into finite discrete distributions by computing their similarities to a number of predefined anchor points in the data space. Then, the Fisher information distance on the statistical manifold is used as the distance in the input feature space. This induces a new family of Riemannian distance metric in the input data space with two important properties. First, the new
Riemannian metric is robust to density variation in the original data space. Without such robustness, an objective function can be easily biased towards data regions the density of which is low and thus dominates learning of the objective function. Second, the new Riemannian metric has largest distance discrimination on the manifold of anchor points and no distance in the directions being orthogonal to the manifold. So, the effect of locally irrelevant dimensions of anchor points is removed. To the best of our knowledge, this is the first metric learning algorithm that has these two important properties.

SBFIML is flexible and general; it can be applied to different types of data spaces with various non-negative similarity functions. Comparing to KML, SBFIML does not require the similarity measure to form a PSD matrix. Moreover, SBFIML can be interpreted as a local metric learning algorithm. Compared to the previous local metric learning algorithms which produce a non-metric distance~\cite{nohgenerative,junwang2012}, the distance approximation in SBFIML is a well defined distance function with a closed form expression. We evaluate SBFIML on a number of datasets. 
The experimental results show that it outperforms in a statistically significant manner both metric learning methods and SVM.

\section{Preliminaries}
\label{sec:preliminaries}

We are given a number of learning instances $\{\vect x_1, \ldots, \vect x_n\}$, where each instance $\vect x_i^T \in \mathcal X$ is a $d$-dimensional vector, and a vector of associated class labels $\vect y=(y_1,\ldots, y_n)^T$, $y_i \in \{1,\ldots,c\}$. We assume that the input feature space $\mathcal X$ is a smooth manifold. Different learning problems can have very different types of data manifolds with possibly different dimensionality. The most commonly used manifold in metric learning is the Euclidean space $\mathbb R^d$~\cite{weinberger2009distance}. The probability simplex space $\mathcal P^{d-1}$ has also been explored~\cite{lebanon2006metric,cuturi2011ground,Dor2012}. 

We propose a general two-stage metric learning algorithm which can learn a flexible distance in different types of $\mathcal X$ data manifolds, e.g. Euclidean, probability simplex, hypersphere, etc. Concretely, we first map instances from $\mathcal X$ onto the statistical manifold $\mathcal S$ through a similarity-based differential map, which computes their non-negative similarities to a number of predefined anchor points. Then we define the Fisher information distance as the distance on $\mathcal X$. We have chosen to do so, since this induces a new family of Riemannian distance metric which enjoys interesting properties: 1) The new Riemannian metric is robust to density variations in the original data space, which can be produced for example by different intrinsic variabilities of the learning instances in the different categories. Distance learning over this new metric is hence robust to density variation. 2) The new Riemannian distance metric has largest distance discrimination on the manifold of the anchor points and has no distance in the directions being orthogonal to that manifold. So, the new distance metric can remove the effect of locally irrelevant dimensions of the anchor point manifold, see Figure \ref{visulization} for more detials. In the remainder of this section, we will briefly introduce the necessary terminology and concepts. More details can be found in the monographs~\cite{lee2002introduction,amari2007methods}.

\textbf{Statistical Manifold.}
We denote by $\mathcal M^n$ a $n$-dimensional smooth manifold. For each point $p$ on $\mathcal M^n$, there exists at least one smooth coordinate chart $(\mathcal U,\varphi)$ which defines a coordinate system to points on $\mathcal U$, where $\mathcal U$ is an open subset of $\mathcal M^n$ containing $p$ and $\varphi :\mathcal U \longrightarrow \Theta$ is a smooth coordinate map $\varphi(p)= \vect \theta \in \Theta \subset \mathbb{R}^n$. $\vect \theta$ is the coordinate of $p$ defined by $\varphi$. 

A statistical manifold is a smooth manifold whose points are probability distributions. 
Given a $n$-dimensional statistical manifold $\mathcal S^n$, 
we denote by $p(\vect \xi|\vect \theta)$ a probability distribution in $\mathcal S^n$, where $\vect \theta=(\theta_1,\ldots,\theta_n) \in  \Theta \subset \mathbb {R}^n$ is the coordinate of $p(\vect \xi|\vect \theta)$ under some coordinate map $\varphi$ and $\vect \xi$ is the random variable of the $p(\vect \xi|\vect \theta)$ distribution taking values from some set $\Xi$. Note that, all the probability distributions in $\mathcal S^n$ share the same set $\Xi$.

In this paper, we are particularly interested in the $n$-dimensional statistical manifold $\mathcal P^{n}$, whose points are finite discrete distributions,
denoted by
\begin{eqnarray}
\label{m-affine-coordinate}
\mathcal P^{n}=\{p( \xi|\vect \theta=(\theta_1,\ldots,\theta_n)): \sum^n_{i=1}\theta_i<1, \forall i, \theta_i>0 \}
\end{eqnarray}
where $\xi$ is the discrete random variable taking values in the set $\Xi=\{1,\ldots,n+1\}$ and $\vect \theta  \in \Theta \subset \mathbb {R}^{n}$ is called the m-affine coordinate~\cite{amari2007methods}. The probability mass of $p(\xi|\vect \theta)$ is $p(\xi=i)=\theta_i$ if $i \neq n+1$, otherwise $p(\xi=n+1)=1-\sum^n_{k=1}\theta_k$. 

\textbf{Fisher Information Metric.}
The Fisher information metric is a Riemannian metric defined on statistical manifolds and endows a distance between probability distributions~\cite{radhakrishna1945information}. The explicit form of the Fisher information metric at $p(\vect \xi|\vect \theta)$ is a $n \times n$ positive definite symmetric matrix $\mathbf {G}_{FIM}(\vect \theta)$, the $(i,j)$ element of which is defined by:
\begin{eqnarray}
\label{eq:def-FIM}
\mathbf {G}_{FIM}^{ij}(\vect \theta)=\int_{\Xi} \frac{\partial \log p(\vect \xi|\vect \theta)}{\partial \theta_i}\frac{\partial \log p(\vect \xi|\vect \theta)}{\partial \theta_j} p(\vect \xi|\vect \theta) d\vect \xi
\end{eqnarray}
where the above integral is replaced with a sum if $\Xi$ is discrete. 
The following lemma gives the explicit form of the Fisher information metric on $\mathcal P^{n}$.
\newtheorem{lemma}{Lemma}
\begin{lemma}
\label{FIM-Form}
On the statistical manifold $\mathcal P^{n}$, the Fisher information metric $\mathbf {G}_{FIM}(\vect \theta)$ at $p(\xi|\vect \theta)$ with coordinate $\vect \theta$ is 
\begin{eqnarray}
\label{eq:FIM-Form-multinomial}
\mathbf {G}_{FIM}^{ij}(\vect \theta)=\frac{1}{\theta_i}\delta_{ij}+\frac{1}{1-\sum^n_{k=1} \theta_k}, \forall i,j \in \{1,\ldots, n\}
\end{eqnarray}
where $\delta_{ij} =1$ if $i=j$, otherwise $\delta_{ij} =0$.
\end{lemma}

\textbf{Properties of Fisher Information Metric.}
The Fisher information metric enjoys a number of interesting properties. 
First, the Fisher information metric is the unique Riemannian metric induced by all $f$-divergence measures, such as the Kullback-Leibler (KL) divergence and the $\chi^2$ divergence~\cite{amari2010information}. All these divergences converge to the Fisher information distance as the two probability distributions are approaching each other. 
Another important property of the Fisher information metric from a metric learning perspective is that the distance it endows can be approximated by the Hellinger distance, the cosine distance and all $f$-divergence measures~\cite{kass2011geometrical}. More importantly, when $\mathcal S^n$ is the statistical manifold of finite discrete distributions, e.g. $\mathcal P^n$, the cosine distance is exactly equivalent to the Fisher information distance~\cite{lebanon2006metric,lee2007dimensionality}.

\textbf{Pullback Metric.}
Let $\mathcal M^n$ and $\mathcal N^m$ be two smooth manifolds and $\mathcal T_p\mathcal M^n$ be the tangent space of $\mathcal M^n$ at $p \in \mathcal M^n$.
Given a differential map $f:\mathcal M^n \longrightarrow \mathcal N^m$ and a Riemannian metric $\mathbf G$ on $\mathcal N^m$, the differential map $f$ induces a pullback metric $\mathbf G^*$ at each point $ p$ on $\mathcal M^n$ defined by:
\begin{eqnarray}
\label{eq:PullBack-Def}
\left\langle \vect v_1,\vect v_2 \right\rangle_{\mathbf G^*( p)}=\left\langle D_pf( \vect v_1),D_pf( \vect v_2)\right\rangle_ {\mathbf G(f( p))} 
\end{eqnarray}
where $D_pf:\mathcal {T}_{ p} \mathcal M^n \longrightarrow \mathcal {T}_{f( p)} \mathcal N^{m}$ is the differential of $f$ at point $p \in \mathcal M^n$, which maps tangent vectors $\vect v \in \mathcal {T}_{ p} \mathcal M^n$ to tangent vectors $D_pf(\vect v) \in \mathcal {T}_{f( p)} \mathcal N^{m}$.

Given the coordinate systems $\Theta$ and $\Gamma$ of $\mathcal U \subset \mathcal M^n$ and $\mathcal U' \subset \mathcal N^m$ respectively, 
defined by some smooth coordinate maps $\varphi_{\mathcal U}$ and  $\varphi_{\mathcal U'}$ 
respectively, then the  explicit form of the pullback metric at point $p \in \mathcal U \subset \mathcal M^n$ with 
coordinate $\vect \theta=\varphi_{\mathcal U}(p)$ is:
\begin{eqnarray}
\label{eq:PullBack-Form}
\mathbf {G^*}(\vect \theta)=\mathbf J^T \mathbf G(\vect \gamma) \mathbf J
\end{eqnarray}
where $\vect \gamma=\varphi_{\mathcal U'}(f(p))$ is the coordinate of the $f(p) \in \mathcal U' \subset \mathcal N^m$
and $\mathbf J$ is the Jacobian matrix of the function 
$\varphi_{\mathcal U'}\circ f \circ\varphi^{-1}_{\mathcal U}: \Theta \longrightarrow \Gamma$ 
at point $\vect \theta$. Since $\mathbf G$ is a Riemannian metric, the pullback metric $\mathbf G^*$ is in general at least 
a PSD metric. 

The following lemma gives the relation between the geodesic distances on $\mathcal M^n$ and $\mathcal N^m$.
\begin{lemma}
\label{pullback-property}
Let $\mathbf G^*$ be the pullback metric of a Riemannian metric $\mathbf G$ induced by a differential map $f:\mathcal M^n \longrightarrow \mathcal N^m$,
$d_{\mathbf G^*}(p',p )$ be the geodesic distance on $\mathcal M^n$ endowed by $\mathbf G^*$ and $d_{\mathbf G}(f(p'),f(p))$ the geodesic distance on $\mathcal N^m$ endowed by $\mathbf G$, then, it holds
$\lim_{p' \to p}\frac{d_{\mathbf G} (f(p'),f(p ))}{d_{\mathbf G^*}(p',p )}=1$
\end{lemma}
The proof of Lemma \ref{pullback-property} is provided in the appendix. 
In addition to approximating $d_{\mathbf G^*}(p',p)$ directly on $\mathcal M^n$ by assuming that the geodesic curve is formed by straight lines as previous local metric learning algorithms do~\cite{nohgenerative,junwang2012}, Lemma \ref{pullback-property} allows us to also approximate it with $d_{\mathbf G} (f(p'),f(p ))$ on $\mathcal N^m$. Note that, both approximations have the same asymptotic convergence result.

\section{Similarity-Based Fisher Information Metric Learning}
\label{sec:fiml}
We will now present our two-stage metric learning algorithm, SBFIML. 
In the following, we will first present how to define the similarity-based differential map $f: \mathcal X \longrightarrow  \mathcal P$ and then how to learn the Fisher information distance.

\subsection{Similarity-Based Differential Map}

Given a number of anchor points $\{\vect z_1,\ldots,\vect z_n\}$, $\vect z_i \in \mathcal X$, we denote by 
$s=(s_1,\ldots,s_n):\mathcal X \longrightarrow \mathbb {R^+}^n$ the differentiable similarity function. Each 
$s_k:\mathcal X \longrightarrow \mathbb {R^+}$ component is a differentiable function the output of which
is a non-negative similarity between some input instance $\vect x_i$ and the anchor point $\vect z_k$.
Based on the similarity function $s$ we define the similarity-based differential map $f$ as:
\begin{eqnarray}
\label{eq:DifferentialMap}
f(\vect x_i) &&= p(\xi|(\frac{s_1(\vect x_i)}{\sum^n_{k=1} s_k(\vect x_i)},\ldots,\frac{s_{n-1}(\vect x_i)}{\sum^n_{k=1} s_k(\vect x_i)}))\\\nonumber
&&=({\bar{s}_1(\vect x_i)},\ldots,{\bar{s}_{n-1}(\vect x_i)})
\end{eqnarray}
where $f(\vect x_i)$ is a finite discrete distribution on manifold $\mathcal P^{n-1}$. From now on, for simplicity, we will denote $f(\vect x_i)$ by $p^i(\xi)$. The probability mass of the $k$th outcome is given by:
$p^i(\xi=k)={\bar{s}_k(\vect x_i)=\frac{s_k(\vect x_i)}{\sum^n_{k=1} s_k(\vect x_i)}}$. 
In order for $f$ to be a valid differential map, the similarity function $s$ must satisfy $\sum_k s_k(\vect x_i)>0, \ \forall \vect x_i \in \mathcal X$. 
This family of differential maps is very general and can be applied to any $\mathcal X$ space where a non-negative differentiable similarity 
function $s$ can be defined. 
The finite discrete distribution representation, $p^i(\xi)$, of learning instance, $\vect x_i$, can be intuitively seen as an encoding of its neighborhood structure defined by the similarity function $s$. Note that, the idea of mapping instances onto the statistical manifold $\mathcal P$ has been previously studied in manifold learning, e.g. SNE~\cite{hinton2002stochastic} and t-SNE~\cite{van2008visualizing}. 

Akin to the appropriate choice of the kernel function in a kernel-based method, 
the choice of an appropriate similarity function $s$ is also crucial for SBFIML. 
In principle, an appropriate similarity function $s$ should be a good match for the geometrical 
structure of the $\mathcal X$ data manifold.  For example, for data lying on the probability 
simplex space, i.e. $\mathcal X=\mathcal P^{d-1}$, the similarity functions defined either on $\mathbb R^d$ 
or on $\mathcal P^{d-1}$ can be used. However, the similarity 
function on  $\mathcal P^{d-1}$ is more appropriate, because it exploits the geometrical structure of $\mathcal P^{d-1}$, which, in contrast, is 
ignored by the similarity function on $\mathbb R^d$~\cite{Dor2012}.
\note[Alexandros]{Now you confused me. You talk an appropriate similarity function on $\mathcal X$ 
and you bring the example of the probability simplex space. However later we define an appropriate 
distance over the probability simplex $\mathcal S$ induced by $f(\vect x_i)$, which never the less 
is different from $\mathcal X$. It could confuse the reader.}

The set of anchor points $\{\vect z_1,\ldots,\vect z_n\}$ can be defined in various ways. Ideally, anchor points should be similar to the given learning instances $\vect x_i$, i.e. anchor points follow the same distribution as that of learning instances. Empirically, we can use 
directly training instances or cluster centers, the latter established by clustering algorithms. 
Similar to the current practice in kernel methods we will use in SBFIML as anchors points all the training instances. 

\begin{figure*}[t]
  \centering
	\subfigure[$\mathbf G_{\mathcal Q}^*(\vect x)$]{\label{kernel-line}\includegraphics[width=0.23\textwidth]{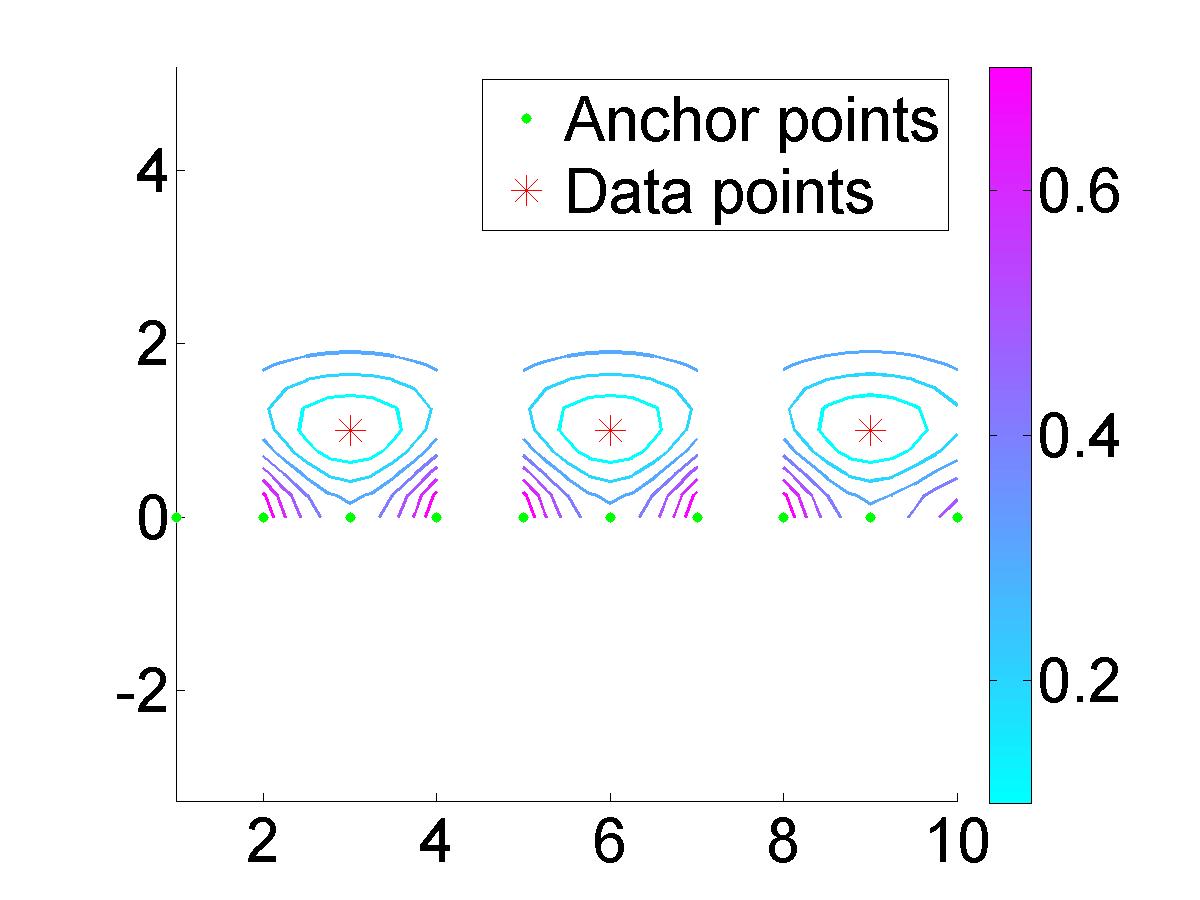}}
\subfigure[$\mathbf G_{\mathcal P}^*(\vect x)$]{\label{fisher-line}\includegraphics[width=0.23\textwidth]{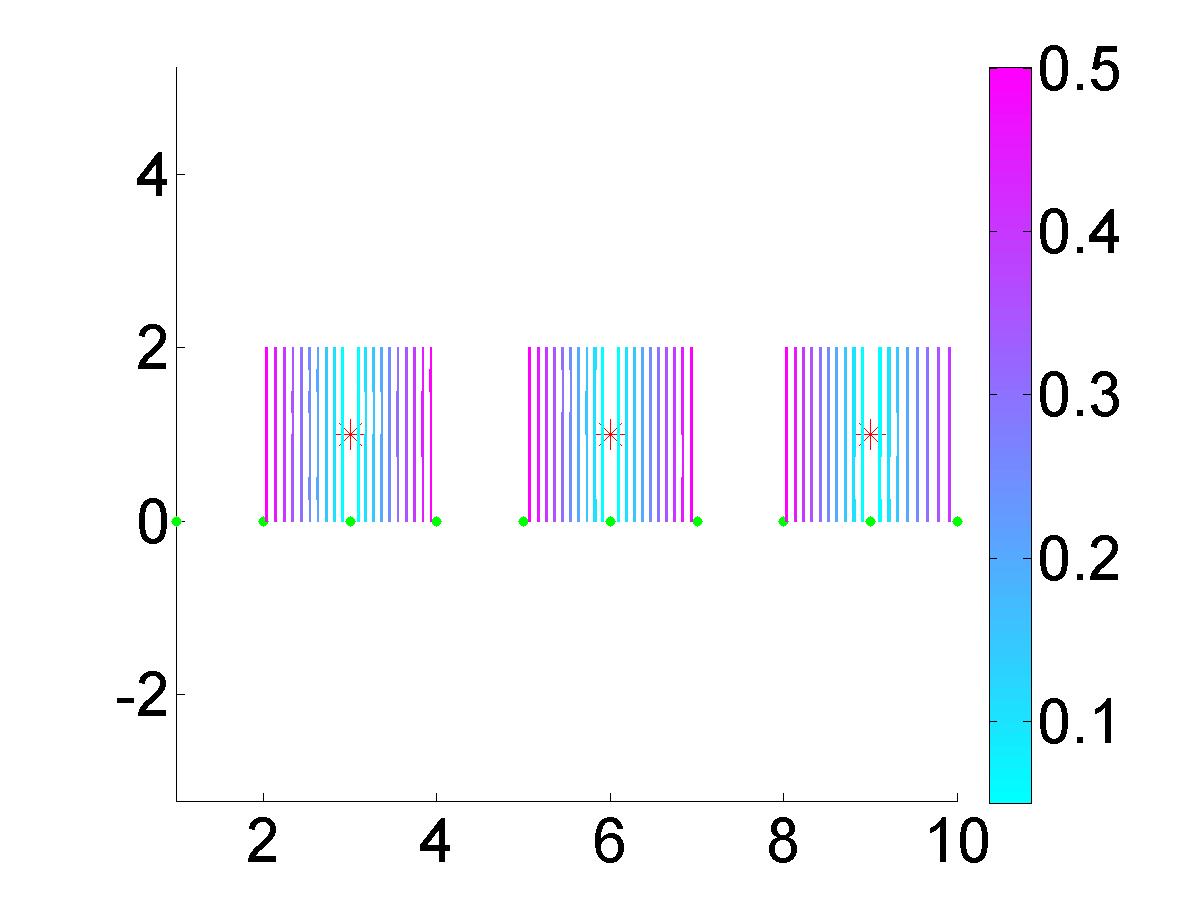}}
\centering
  \subfigure[$\mathbf G_{\mathcal Q}^*(\vect x)$]{\label{kernel-manifold}\includegraphics[width=0.23\textwidth]{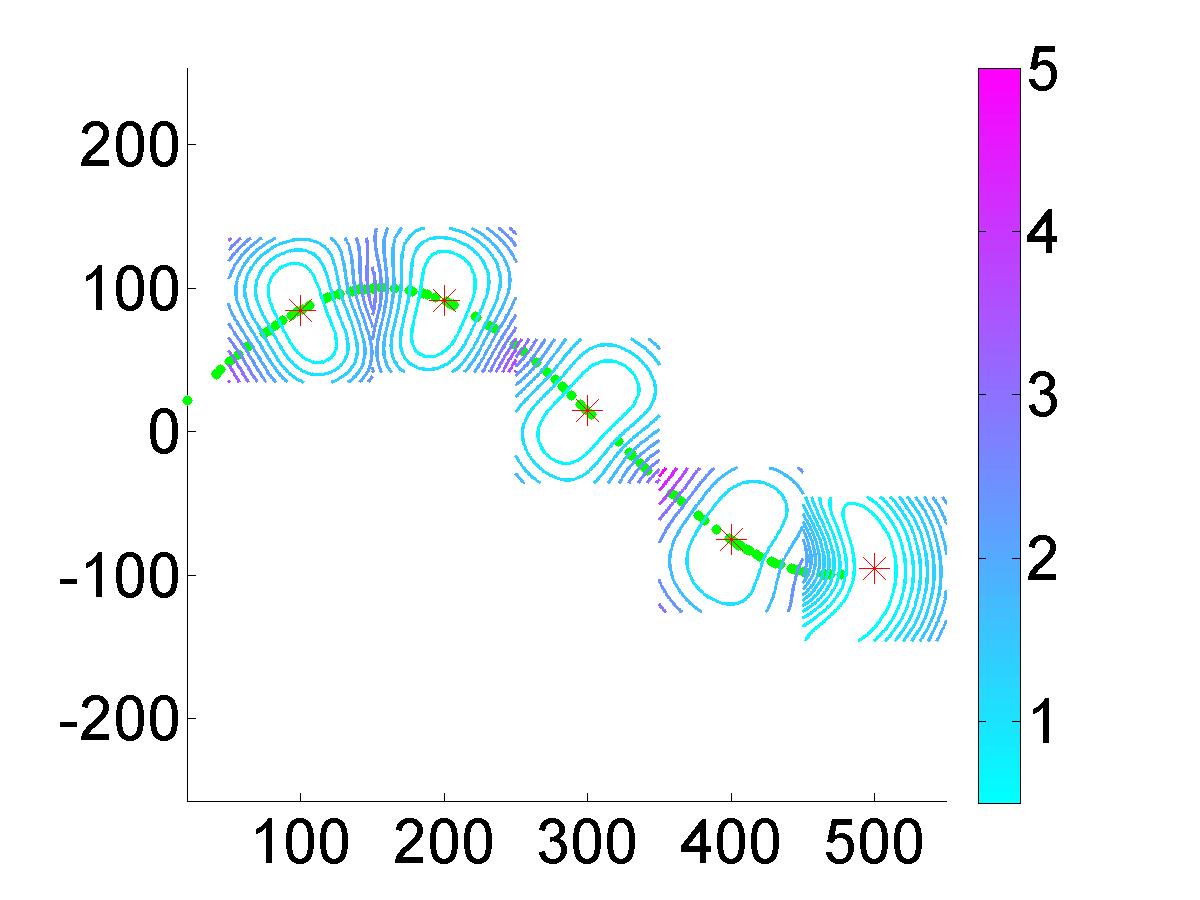}}
  \subfigure[$\mathbf G_{\mathcal P}^*(\vect x)$]{\label{fisher-manifold}\includegraphics[width=0.23\textwidth]{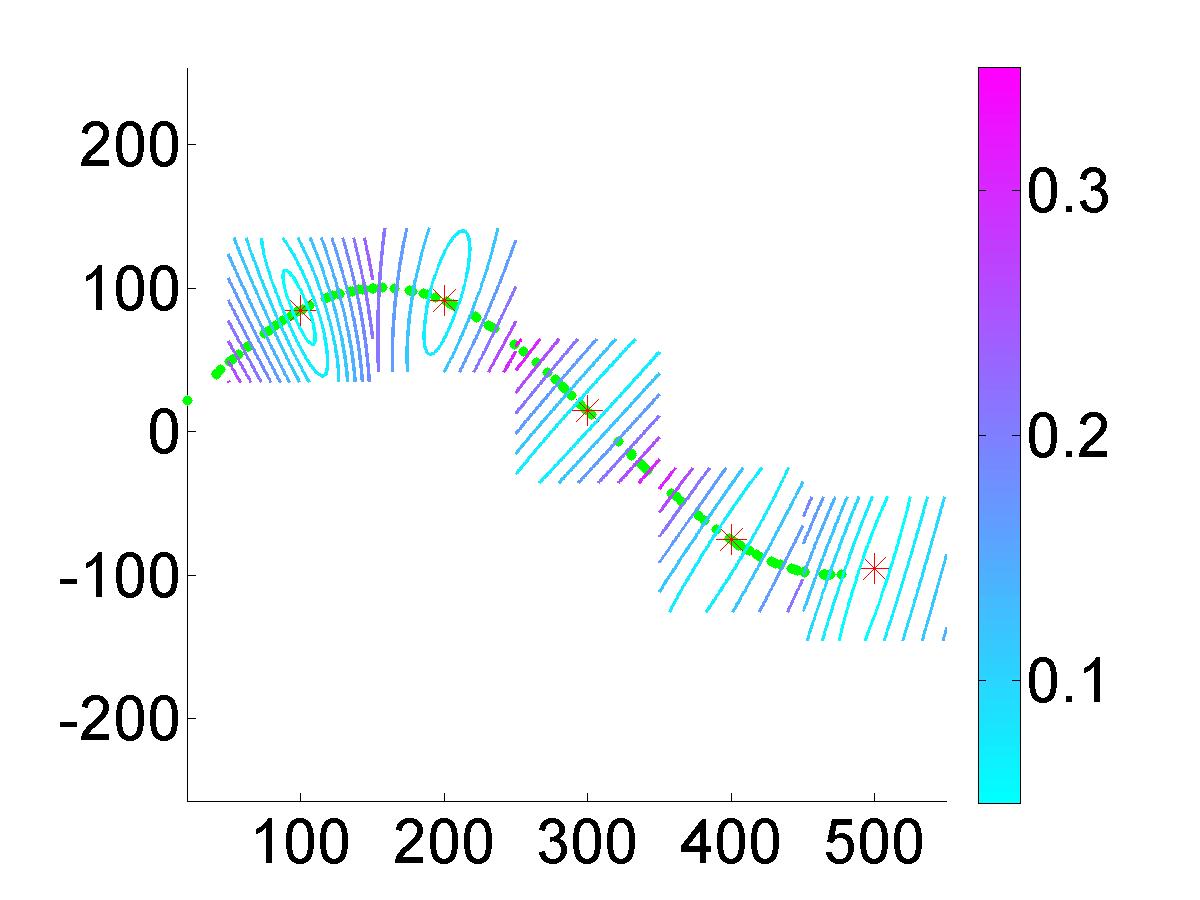}}  
  \caption{The visulization of equi-distance curves of pullback metrics $\mathbf G_{\mathcal Q}^*(\vect x)$ and $\mathbf G_{\mathcal P}^*(\vect x)$. }
\label{visulization}
\end{figure*}

\textbf{Similarity Functions on $\mathbb R^d$.} We can define the similarity on $\mathbb R^d$ in various ways. 
In this paper we will investigate two types of differentiable similarity functions. The first one is based on the Gaussian function, defined as: 
\begin{eqnarray}
\label{eq:Similarity_gaussian}
s_k(\vect x_i)= exp(-\frac{\|\vect x_i-\vect z_k \|^2_2}{\sigma_k})
\end{eqnarray}
where $\| \cdot \|_2$ is the $L_2$ norm. $\sigma_k$ controls the size of the neighborhood of the anchor 
point $\vect z_k$, with large values producing large neighborhoods. Note that the different $\sigma_k$s
could be set to different values; if all of them are equal, this similarity function is exactly 
the Gaussian kernel. 
The second type of similarity function that we will look at is:
\begin{eqnarray}
\label{eq:Similarity_Cosine}
s_k(\vect x_i)= 1-\frac{1}{\pi}\arccos(\frac{\vect x^T_i\vect z_k }{\|\vect x_i \|_2 \cdot \|\vect z_k \|_2 })
\end{eqnarray}
which measures the normalized angular similarity between $\vect x_i$ and $\vect z_k$. This similarity function can be explained as we first projecting all points from $\mathbb R^d$ to the hypersphere and then applying the angular similarity to points on a hypersphere. As a result, this similarity function is useful for data which approximately lie on a hypersphere.
Note that this similarity function is also a valid kernel function~\cite{honeine2010angular}. 
\note[Alexandros]{You see? with respect to my previous comment in which you were talking about appropriate distance functions and brough the probability simplex space example
as a case in which we need a special metric. Here you do not use similarities that exploit structure.}

One might say we can also achieve nonlinearity by mapping instances into the proximity space $\mathcal Q$ using the following similarity-based map $g: \mathcal X \longrightarrow  \mathcal Q $:
\begin{eqnarray}
\label{eq:NonNormalizedDifferentialMap}
g(\vect x) = ({s_1(\vect x)},\ldots,{s_{n}(\vect x)})
\end{eqnarray}
We now compare our similarity-based map $f$, equation \ref{eq:DifferentialMap} against the similarity-based map $g$, equation \ref{eq:NonNormalizedDifferentialMap}, in two aspects, namely representation robustness and pullback metric analysis. 

\textbf{Representation Robustness.}
Compared to the representation induced by the similarity-based map $g$, equation \ref{eq:NonNormalizedDifferentialMap}, our representation induced by the similarity-based map $f$, equation \ref{eq:DifferentialMap}, is more robust to density variations in original data space, i.e. the density of the learning instances varies significantly between different regions. This can be explained by the fact that the finite discrete distribution is essentially a representation of the neighborhood structure of a learning instance normalized by a "scaling" factor, the sum of similarities of the learning instance to the anchor points. Hence the distance implied by the finite discrete distribution representation is less sensitive to the density variations of the different data regions. This is an important property. Without such robustness, an objective function based on raw distances can be easily biased towards data regions the density of which is low and thus dominates learning of the objective function. One example of this kind of objective is that of LMNN \cite{weinberger2009distance}, which we will also use later in SBFIML to learn the Fisher information distance.

\textbf{Pullback Metric Analysis.}
We also show how the two approaches differ by comparing the pullback metrics induced by the two similarity-based maps $f$ and $g$. In doing so, we first need to specify the Riemannian metrics $\mathbf G_{\mathcal Q}$ in the proximity space $\mathcal Q$ and $\mathbf G_{\mathcal P}$ on the statistical manifold ${\mathcal P^{n-1}}$. Following the work of similarity-based learning~\cite{Chen2009}, we use the Euclidean metric as the $\mathbf G_{\mathcal Q}$ in the proximity space $\mathcal Q$. On the statistical manifold $\mathcal P^{n-1}$ we use the Fisher information metric $\mathbf G_{FIM}$ defined in equation \ref{eq:FIM-Form-multinomial} as $\mathbf G_{\mathcal P}$. To simplify our analysis, we assume $\mathcal X=\mathbb R^d$. However, note that this analysis can be generalized to other manifolds, e.g. $\mathcal P^{d-1}$.  We use the standard Cartesian coordinate system for points in $\mathbb R^d$ and $\mathcal Q$ and use m-affine coordinate system, equation \ref{m-affine-coordinate}, for points on $\mathcal P^{n-1}$.

The pullback metric induced by these two differential maps are given in the following lemma.
\begin{lemma}
\label{pullback-Form}
In $\mathbb R^d$, at $\vect x$ with Cartesian coordinate, the form of the pullback metric $\mathbf {G_{\mathcal Q}^*}(\vect x)$ of the Euclidean metric induced by the differential map $g$ of equation \ref{eq:NonNormalizedDifferentialMap} is: 
\begin{eqnarray}
\label{eq:pullback-form-nonnormalized}
 \mathbf {G_{\mathcal Q}^*}(\vect x)=\nabla g({\vect x})  \nabla g({\vect x})^T=\sum^n_{i=1} \nabla s_i(\vect x) \nabla s_i(\vect x)^T
\end{eqnarray}
where the vector $\nabla s_i(\vect x)$ of size $d \times 1$ is the differential of $i$th similarity function $s_i(\vect x)$. The form of the pullback metric $\mathbf {G_{\mathcal P}^*}(\vect x)$ of the Fisher information metric induced by the differential map $f$ of equation \ref{eq:DifferentialMap} is:
\begin{eqnarray}
\label{eq:pullback-form-normalized}
 \mathbf {G_{\mathcal P}^*}(\vect x)&&=\sum^n_{i=1} \frac{1}{\bar{s}_i(\vect x)} (\nabla {\bar{s}_i(\vect x)} \nabla {\bar{s}_i(\vect x)}^T)
\end{eqnarray}
where $\nabla \bar{s}_i(\vect x)={\bar{s}_i(\vect x)} \left(\nabla \log(s_i(\vect x)) - E\left( \nabla \log(s_i(\vect x))\right)\right)$ and the expectation of $\nabla \log(s_i(\vect x))$ is $ E( \nabla \log(s_i(\vect x)))=\sum^n_{k=1}  {\bar{s}_k(\vect x)} \nabla \log(s_i(\vect x))$ .
\end{lemma}

\textbf{Gaussian Similarity Function.}
The form of pullback metrics $\mathbf {G_{\mathcal Q}^*}(\vect x)$ and $\mathbf {G_{\mathcal P}^*}(\vect x)$ depends on the explicit form of the similarity function $s_i(\vect x)$. We now study their differences using the Gaussian similarity function with kernel width $\sigma$, equation \ref{eq:Similarity_gaussian}.
We first show the difference between $\mathbf {G_{\mathcal Q}^*}(\vect x)$ and $\mathbf {G_{\mathcal P}^*}(\vect x)$ by comparing their $m$ largest eigenvectors, the directions in which metrics have the largest distance discrimination.

The $m$ largest eigenvectors $\mathbf U_{\mathcal Q}(\vect x)$ of $\mathbf {G_{\mathcal Q}^*}(\vect x)$ are:
\begin{eqnarray}
\label{discriminative-nonnormalized}
\mathbf U_{\mathcal Q}(\vect x) &&= \argmax_{\mathbf U^T \mathbf U=\mathbf I}tr(\mathbf {U}^T\mathbf {G_{\mathcal Q}^*}(\vect x) \mathbf {U})\\\nonumber
&&=\argmax_{\mathbf U^T \mathbf U=\mathbf I}\sum^m_{k=1}\sum^n_{i=1}\frac{4}{\sigma^2} (\vect u_k^T s_i(\vect x) (\vect x-\vect z_i))^2
\end{eqnarray}
where $tr(\cdot)$ is the trace norm and $\vect u_k$ is the $k$th column of matrix $\mathbf U$.
The $m$ largest eigenvectors $\mathbf U_{\mathcal P}(\vect x)$ of the pullback metric $\mathbf {G_{\mathcal P}^*}(\vect x)$ are:
\begin{eqnarray}
\label{discriminative-normalized}
\mathbf U_{\mathcal P}(\vect x)&&=
 \argmax_{\mathbf U^T \mathbf U=\mathbf I}tr(\mathbf {U}^T\mathbf {G_{\mathcal P}^*}(\vect x) \mathbf {U})\\\nonumber
&&=\argmax_{\mathbf U^T \mathbf U=\mathbf I}\sum^m_{k=1}\sum^n_{i=1} \frac{4\bar{s}_i(\vect x)}{\sigma^2} (\vect u_k^T(\vect z_i-E(\vect z_i))^2
\end{eqnarray}
where $E(\vect z_i)=\sum^n_{k=1} {\bar{s}_k(\vect x)} \vect z_k$


We see one key difference between $\mathbf U_{\mathcal P}(\vect x)$ and $\mathbf U_{\mathcal Q}(\vect x)$. In equation \ref{discriminative-normalized}, $\mathbf U_{\mathcal P}(\vect x)$ are the directions which maximize the sum of expected variance of $\vect u_k^T\vect z_i, k \in\{1,\ldots,m\},$ with respected to its expected mean. In contrast, the directions of $\mathbf U_{\mathcal Q}(\vect x)$ in equation \ref{discriminative-nonnormalized} maximize the sum of the unweighted "variance" of $\vect u_k^T s_i(\vect x) (\vect x-\vect z_i), k \in\{1,\ldots,m\},$ without centralization. Their difference can be intuitively compared to the difference of doing local PCA with or without centralization. Therefore, $\mathbf U_{\mathcal P}(\vect x)$ is closer to the principle directions of local anchor points. 
Second, since 
$\mathbf {G_{\mathcal P}^*}(\vect x)=\sum^n_{i=1} \frac{4\bar{s}_i(\vect x)}{\sigma^2} (\vect z_i-E(\vect z_i))(\vect z_i-E(\vect z_i))^T,$
it is also easy to show that $\mathbf {G_{\mathcal P}^*}(\vect x)$ has no distance in the orthogonal directions of the affine subspace spanned by the weighted anchor points of $\bar{s_i}(\vect x) \vect z_i$. So, $\mathbf {G_{\mathcal P}^*}(\vect x)$ removes the effect of locally irrelevant dimensions to the anchor point manifold. 


To show the differences of pullback metrics $\mathbf G_{\mathcal Q}^*(\vect x)$ and $\mathbf G_{\mathcal P}^*(\vect x)$ intuitively, we visualize their equi-distance curves in Figure \ref{visulization}, where the Guassian similarity function, euqation \ref{eq:Similarity_gaussian}, is used to define the similarity maps in equations \ref{eq:NonNormalizedDifferentialMap} and \ref{eq:DifferentialMap}. As shown in Figure \ref{visulization}, we see that the pullback metric $\mathbf G_{\mathcal P}^*(\vect x)$ emphasizes more the distance along the principle direction of the local anchor points than the pullback metric $\mathbf G_{\mathcal Q}^*(\vect x)$. Furthermore, in Figure \ref{fisher-line} we see that $\mathbf G_{\mathcal P}^*(\vect x)$ has a zero distance in the direction being orthogonal to the manifold of anchor points, the straight line which the (green) anchor points lie on. Therefore, $\mathbf G_{\mathcal P}^*(\vect x)$ is more discriminative on the manifold of the anchor points.
To explore the effect of these differences, we also experimentally compare these two approaches in section \ref{sec:exp} and the results show that learning the Fisher information distance on $\mathcal P$ outperforms in a significant manner learning Mahalanobis distance in proximity space $\mathcal Q$.


\subsection{Large Margin Fisher Information Metric Learning}
By applying on the learning instances the differential map $f$ of equation (\ref{eq:DifferentialMap}) we map them on the statistical manifold 
$\mathcal P^{n-1}$. We are now ready to learn the Fisher information distance from the data. 

\textbf{Distance Parametrization.}
As discussed in section \ref{sec:preliminaries}, the Fisher information distance on $\mathcal P^{n-1}$ can be exactly computed by the cosine distance~\cite{lebanon2006metric,lee2007dimensionality}:
\begin{eqnarray}
\label{dis_FIM}
d_{FIM}(\vect p^i,\vect p^j)=2\arccos (\sqrt{{\vect p^i}}^T \sqrt{\vect p^j})
\end{eqnarray}
where $\vect p^i$ is the probability mass vector of the finite discrete distribution $p^i(\xi)$. To parametrize the Fisher information distance, we apply on the probability mass vector  $\vect p^i$ a linear transformation $\mathbf L$. The intuition is that,  the effect of the optimal linear transformation $\mathbf L$ is equivalent to locating a set of hidden anchor points such that the data's similarity representation is the same as the transformed representation. Thus the parametric Fisher information distance is defined as:
\begin{eqnarray}
\label{dis_FIM_parmetrized}
 d_{FIM}(\mathbf L \vect p^i,\mathbf L \vect p^j)=&&2\arccos (\sqrt{{\mathbf L \vect p^i}}^T \sqrt{\mathbf L \vect p^j}) \\\nonumber
 s.t. &&  {\mathbf L \geq 0}, \sum_i L_{ij} =1, \forall j 
\end{eqnarray}
$\mathbf L$ has size $k \times n$. $k$ is the number of hidden anchor points. To speedup the learning process, in practice we often learn a low rank linear 
transformation matrix $\mathbf L$ with small $k$. The constraints ${\mathbf L \geq 0}$ and $ \sum_i L_{ij} =1, \forall j$ are added to ensure 
that each ${\mathbf L} \vect p^i$ is still a finite discrete distribution on the manifold $\mathcal P^{k-1}$. 

\textbf{Learning.}
We will follow the large margin metric learning approach of~\citep{weinberger2009distance} and define the optimization problem of learning $\mathbf L$ as:
\begin{eqnarray}
\label{opt:FIML}
\min_{\mathbf L} && \sum_{ijk \in C(i, j, k)} [\epsilon_{ijk}]_+ +\alpha \sum_{i,j \rightarrow i} d_{FIM}(\mathbf L\vect p^i,\mathbf L\vect p^j)\\\nonumber
s.t. 
&& \mathbf L \geq 0 \\\nonumber
&& \sum_{i} L_{ij} =1 ; \ \forall { j} \\\nonumber
&&\epsilon_{ijk}=d_{FIM}(\mathbf L\vect p^i,\mathbf L\vect p^j)+\gamma -d_{FIM}(\mathbf L\vect p^i,\mathbf L\vect p^k)\\\nonumber
\end{eqnarray}
where $\alpha$ is a parameter that balances the importance of the two terms. Unlike LMNN~\cite{weinberger2009distance}, 
the margin parameter $\gamma$ is added in the large margin triplet constraints following the work of~\cite{Dor2012}, since the cosine distance is not linear with $\mathbf {L^TL}$. 
The large margin triplet constraints $C(i,j,k)$ for each instance $\vect x_i$ are generated using its $k_1$ same-class nearest neighbors and its $k_2$ different-class nearest neighbors 
in the $\mathcal X$ space and constraining the distance of each instance to its $k_2$ different class 
neighbors to be larger than those to its $k_1$ same class neighbors with $\gamma$ margin.  In the objective 
function of (\ref{opt:FIML}) the matrix $\mathbf L$ is learned by minimizing the sum of the hinge losses and 
the sum of the pairwise distances of each instance to its $k_1$ same-class nearest neighbors. 

\textbf{Optimization}. Since the cosine distance defined in equation (\ref{dis_FIM}) is not convex, the optimization problem (\ref{opt:FIML}) 
is not convex. However, the constraints on matrix $\mathbf L$ are linear and we can solve this problem using a projected sub-gradient 
method. At each iteration, the main computation is the sub-gradient computation with complexity $O(mnk)$, where $m$ is the number of 
large margin triplet constraints. $n$ and $k$ are the dimensions of the $\mathbf L$ matrix. The simplex projection operator on matrix $\mathbf L$ can be 
efficiently computed with complexity $O(nk\log(k))$~\cite{duchi2008efficient}. 
Note that, learning distance metric on $\mathcal P$ has been previously studied by Riemannian Metric Learning (RML)~\cite{lebanon2006metric} and $\chi^2$-LMNN~\cite{Dor2012}. In $\chi^2$-LMNN, a symmetric $\chi^2$ distance on $\mathcal P$ is learned with large margin idea similar to problem \ref{opt:FIML}. SBFIML differs from $\chi^2$-LMNN in that it uses the cosine distance to measure the distance on $\mathcal P$. As described in section \ref{sec:preliminaries}, the cosine distance is exactly equivalent to the Fisher information distance on $\mathcal P$, while the $\chi^2$ distance is only an approximation. In contrast to SBFIML and $\chi^2$-LMNN, the work of RML focuses on unsupervised Fisher information metric learning. More importantly, both RML and $\chi^2$-LMNN can only be applied in problems in which the input data lie on $\mathcal P$, while SBFIML can be applied to general data manifolds via the similarity-based differential map. Finally, note that SBFIML can also be applied to problems where we only have access to the pairwise instance similarity matrix, since it needs only the probability mass of finite discrete distributions as its input. 


\textbf{Local Metric Learning View of SBFIML.}
SBFIML can also be interpreted as a local metric learning algorithm. SBFIML defines the local metric on $\mathcal X$ as the pullback metric of the Fisher information metric induced by the following similarity-based parametric differential map $f_{\mathbf L}: \mathcal X  \longrightarrow  \mathcal P^{k-1}$:
\begin{eqnarray}
\label{eq:ParametricMap}
f_{\mathbf L}(\vect x_i)= {\mathbf L} \cdot \vect p^i,  s.t. & {\mathbf L >0}, \sum_i L_{ij} =1, \forall j 
\end{eqnarray}
where as before $\vect p^i $ is the probability mass vector of the finite discrete distribution $p^i(\xi)$ defined in equation (\ref{eq:DifferentialMap}). 
SBFIML learns the local metric by learning the parameters of $f_{\mathbf L}$.
The explicit form of the pullback metric $\mathbf G^*$ can be computed according to the equation (\ref{eq:PullBack-Form}).
Given the pullback metric we can approximate the geodesic distance on $\mathcal X$ by assuming that the geodesic curves are formed by straight lines as local metric learning methods~\cite{nohgenerative,junwang2012} do, which would result in a non-metric distance. However, Lemma \ref{pullback-property} allows us to approximate the geodesic distance on $\mathcal X$ by the Fisher information distance on $\mathcal P^{k-1}$. SBFIML follows the latter approach. Compared to the non-metric distance approximation, this new distance is a well defined distance function which has a closed form expression. Furthermore, this new distance approximation has the same asymptotic convergence result as the non-metric distance approximation.

\begin{table*}[bt!]
\begin{center}
\caption{Mean and standard deviation of 5 times 10-fold CV accuracy results on $\mathbb R^d$ datasets. The superscripts $^{+-=}$ next to the accuracies of SBFIML 
indicate the result of the Student's t-test with SBMML,$\chi^2$ LMNN, LMNN, GLML, PLML, KML and SVM. They denote respectively a significant 
win, loss or no difference for SBFIML. The \textbf{bold} entries for each dataset have no significant difference from the best accuracy
for that dataset. The number in the parenthesis indicates the score of the respective algorithm for the given dataset based 
on the pairwise comparisons of the Student's t-test.}
\label{results-Euclidean}
\vskip 0.15in
 \scalebox{0.6}{
\begin{tabular}{l||c||c||c||c||c||c||c||c}
          
Datasets(\#Inst./\#Feat./\#Class)    & SBFIML    &SBMML    &$\chi^2$ LMNN                    &LMNN                &GLML   &PLML     &KML  &SVM\\ \hline \hline
stk25(208/172/2)   &\textbf{  81.6$\pm$1.8}$^{==+++==}$(5.0)&\textbf{  81.7$\pm$3.0}(5.0)&\textbf{  80.9$\pm$1.4}(5.0)&  75.7$\pm$2.0$$(1.5)&  72.7$\pm$1.8$$(0.5)&  74.4$\pm$3.6$$(1.0)&\textbf{  81.9$\pm$2.7}(5.0)&\textbf{  81.2$\pm$1.0}(5.0)\\ 
wpbc(198/33/2)   &\textbf{  79.6$\pm$1.0}$^{==+++=+}$(5.5)&\textbf{  79.3$\pm$1.1}(5.5)&\textbf{  78.8$\pm$1.7}(4.5)&  73.6$\pm$1.7$$(0.5)&  76.5$\pm$1.6$$(3.0)&  71.7$\pm$1.6$$(0.5)&\textbf{  79.7$\pm$1.2}(5.5)&  77.3$\pm$0.6$$(3.0)\\ 
wine(178/13/3)   &\textbf{  98.0$\pm$1.0}$^{===+===}$(4.0)&\textbf{  98.3$\pm$0.4}(5.0)&  97.4$\pm$0.3$$(3.5)&  97.3$\pm$0.5$$(3.5)&  96.1$\pm$1.1$$(0.5)&\textbf{  97.5$\pm$1.1}(3.5)&\textbf{  98.1$\pm$0.6}(4.0)&\textbf{  98.1$\pm$0.6}(4.0)\\ 
sonar(208/60/2)   &\textbf{  87.2$\pm$1.3}$^{==+====}$(4.0)&\textbf{  87.1$\pm$2.0}(3.5)&\textbf{  86.4$\pm$2.0}(3.5)&  84.8$\pm$1.5$$(2.0)&  87.1$\pm$0.7$$(3.5)&  86.1$\pm$1.4$$(3.0)&\textbf{  86.9$\pm$2.2}(3.5)&\textbf{  88.1$\pm$0.2}(5.0)\\ 
musk(476/166/2)   &\textbf{  96.1$\pm$0.4}$^{++=++++}$(6.5)&  95.5$\pm$0.2$$(4.5)&  94.8$\pm$0.5$$(3.0)&\textbf{  95.8$\pm$0.5}(5.0)&  91.3$\pm$0.6$$(0.5)&  90.9$\pm$0.3$$(0.5)&  95.3$\pm$0.2$$(4.0)&  94.9$\pm$0.7$$(4.0)\\ 
wdbc(569/30/2)   &  97.2$\pm$0.4$^{-=++=-=}$(3.5)&\textbf{  97.9$\pm$0.3}(6.0)&\textbf{  97.5$\pm$0.5}(4.5)&  96.4$\pm$0.2$$(1.0)&  96.1$\pm$0.4$$(0.5)&  96.8$\pm$0.5$$(3.0)&\textbf{  97.9$\pm$0.3}(6.0)&  97.3$\pm$0.2$$(3.5)\\ 
balance(625/4/3)   &\textbf{  97.5$\pm$0.5}$^{++++++=}$(6.5)&  96.6$\pm$0.3$$(4.0)&  96.2$\pm$0.5$$(4.0)&  90.2$\pm$0.8$$(1.5)&  88.8$\pm$0.5$$(0.0)&  91.8$\pm$2.0$$(1.5)&  96.6$\pm$0.3$$(4.0)&\textbf{  97.7$\pm$0.5}(6.5)\\ 
breast(683/10/2)   &\textbf{  96.7$\pm$0.3}$^{==+=+==}$(4.5)&\textbf{  96.4$\pm$0.5}(4.0)&\textbf{  96.9$\pm$0.3}(5.0)&  95.8$\pm$0.4$$(1.0)&  96.4$\pm$0.2$$(3.5)&  95.1$\pm$0.7$$(0.5)&\textbf{  96.5$\pm$0.4}(4.5)&\textbf{  96.9$\pm$0.2}(5.0)\\ 
australian(690/14/2)   &  84.6$\pm$0.3$^{++++++-}$(6.0)&  80.5$\pm$0.9$$(2.0)&  83.5$\pm$0.5$$(5.0)&  81.2$\pm$1.0$$(2.0)&  80.5$\pm$0.8$$(2.0)&  80.2$\pm$1.0$$(2.0)&  80.8$\pm$0.6$$(2.0)&\textbf{  85.7$\pm$0.9}(7.0)\\ 
vehicle(846/18/4)   &  79.2$\pm$0.6$^{+==+-+=}$(4.5)&  75.7$\pm$1.1$$(1.0)&  78.4$\pm$1.3$$(4.0)&  79.6$\pm$0.9$$(4.5)&  77.3$\pm$0.8$$(2.5)&\textbf{  81.3$\pm$0.5}(6.5)&  76.1$\pm$1.2$$(1.5)&\textbf{  78.0$\pm$7.3}(3.5)\\ 
\hline \hline 
Total Score  &  50.0  &  40.5  &  42.0  &  22.5  &  16.5  &  22.0  &  40.0  &  46.5  \\ 
\end{tabular}
}
\end{center}
\vskip -0.2in
\end{table*}

\section{Experiments}
\label{sec:exp}
We will evaluate the performance of SBFIML on ten datasets from the UCI Machine Learning and mldata\footnote{http://mldata.org/.} repositories. The details of these datasets are reported in the first column of Table \ref{results-Euclidean}.  All datasets are preprocessed by standardizing the input features. We compare SBFIML against 
three metric learning baseline methods: LMNN~\cite{weinberger2009distance}\footnote{http://www.cse.wustl.edu/$\sim$kilian/code/code.html.}, KML~\cite{wang2011metric}\footnote{http://cui.unige.ch/$\sim$wangjun/.}, GLML~\cite{nohgenerative}, and PLML~\cite{junwang2012}. The former two learn 
a global Mahalanobis metric in the input feature space $\mathbb R^d$ and the RKHS space respectively, and the last two learn smooth local metrics in $\mathbb R^d$. In addition, we also compare SBFIML against Similarity-based Mahalanobis Metric Learning (SBMML) to see the difference of pullback metrics $\mathbf G_{\mathcal Q}^*(\vect x)$, equation \ref{eq:pullback-form-nonnormalized}, and $\mathbf G_{\mathcal P}^*(\vect x)$, equation \ref{eq:pullback-form-normalized}. SBMML learns a global Mahalanobis metric in the proximity space $\mathcal Q$. Similar to SBFIML, the metric is learned by optimizing the problem \ref{opt:FIML}, in which the cosine distance is replaced by Mahalanobis distance. The constraints on $\mathbf L$ in problem \ref{opt:FIML} are also removed. To see the difference between the cosine distance used in SBFIML and the $\chi^2$ distance used in $\chi^2$ LMNN, we compare SBFIML against $\chi^2$ LMNN. Note that, both methods solve exactly the same optimization problem \ref{opt:FIML} but with different distance computations. Finally, we also compare SBFIML against SVM for binary classification problems and against multi-class SVMs for multiclass classification problems. In multi-class SVMs, we use the one-against-all strategy to determine the class label.

KML, SBMML and $\chi^2$ LMNN learn a $n \times n$ PSD matrix and are thus computationally expensive for datasets with large number of instances. To speedup the learning process, 
similar to SBFIML, we can learn a low rank transformation matrix $\mathbf L$ of size $k \times n$. For all methods, KML, SBMML, $\chi^2$ LMNN and SBFMIL, we set $k=0.1n$ in all experiments. The matrix $\mathbf L$ in KML and SBMML was initialized by clipping the $n \times n$ identity matrix into the size of $k \times n$. In a similar manner, in $\chi^2$ LMNN and SBFIML the matrix $\mathbf L$ was initialized by applying on the initialization matrix $\mathbf L$ in KML a simplex projector which ensures the constraints in problem (\ref{opt:FIML}) are satisfied.



The LMNN has one hyper-parameter $\mu$~\cite{weinberger2009distance}. We set it to its default value $\mu=1$.
As in ~\cite{nohgenerative}, GLML uses the Gaussian distribution to model the learning instances of a given class. The hyper-parameters of PLML was set following ~\cite{junwang2012}.
The SBFIML has two hyper-parameters $\alpha$ and  $\gamma$.  Following LMNN~\cite{weinberger2009distance}, we set the $\alpha$ parameter 
to $1$. We select the margin parameter $\gamma$ from $\{0.0001,0.001,0.01,0.1\}$ using a 4-fold inner Cross Validation (CV). 
The selection of an appropriate similarity function 
is crucial for SBFIML. We choose the similarity function with a 4-fold inner CV from the angular 
similarity, equation (\ref{eq:Similarity_Cosine}), and the Gaussian similarity in equation (\ref{eq:Similarity_gaussian}). 
We examine two types of Gaussian similarity. In the first we set all $\sigma_k$ to $\sigma$ which is selected 
from $\{0.5 \tau,\tau,2\tau\}$, $\tau$ was set to the average of all pairwise distances. In the second we set the $\sigma_k$ for each 
anchor point $\vect z_k$ separately; the $\sigma_k$ was set by making the entropy of the conditional distribution $p(\vect x_i|\vect z_k)=\frac{s_k(\vect x_i)}{\sum^n_{i=1} s_k(\vect x_i)}$ equal to $\log(nc)$~\cite{hinton2002stochastic}, where $n$ is the number of training instances and $c$ was selected from $\{0.8,0.9,0.95\}$. 

Since $\chi^2$ LMNN and SBFIML apply different distance parametrizations to solve the same optimization problem, the parameters of $\chi^2$ LMNN are set in exactly the same way as SBFIML, except that the margin parameter $\gamma$ of $\chi^2$ LMNN was selected from $\{10^{-8}, 10^{-6},10^{-4},10^{-2}\}$, because $\chi^2$ LMNN uses the squared $\chi^2$ distance \cite{Dor2012}. The best similarity map for $\chi^2$ LMNN is also selected using a 4-fold inner CV from the same similarity function set as that of SBFIML. 
\note[Alexandros]{in the second case I guess that you mean that for each $z_k$ you have a single, but distinct, $\sigma$, 
which is set by the distributional approach you describe, text needs to be adapted. I would like to have that explained :) 
Ovelall don't you have way too many parameters?}

Akin to SBFIML, the performance of KML and SVM depends heavily on the selection of the kernel. We select automatically the best kernel with a 
4-fold inner CV. The kernels are chosen from the linear, the set of polynomial (degree 2,3 and 4), the angular similarity, equation (\ref{eq:Similarity_Cosine}), and the Gaussian kernels with widths $\{0.5 \tau,\tau,2\tau\}$, as in SBFIML $\tau$ was set to the average of all pairwise distances. In addition, we also select the margin parameter $\gamma$ of KML from $\{0.01,0.1,1,10,100\}$. The $C$ parameter of SVM was selected from $\{0.01,0.1,1,10,100\}$. 
SBMML does not have any constraints on the similarity function, thus we select its similarity function with a 4-fold inner CV from a set which includes all kernel and similarity functions used in SBFIML and KML. As in KML, we select the margin parameter $\gamma$ of SBMML from $\{0.01,0.1,1,10,100\}$.
For all methods, except GLML and SVM which do not involve triplet constraints, the triplet constraints are constructed using three 
same-class and ten different-class nearest neighbors for each learning instance. Finally, we use the 1-NN rule to evaluate the performance of the different metric learning methods. 

To estimate the classification accuracy we used 5 times 10-fold CV. The statistical significance of the differences were tested 
using Student's t-test with a p-value of 0.05. In order to get a better understanding of the relative performance of the different algorithms 
for a given dataset we used a simple ranking schema in which an algorithm A was assigned one point if it was found to have 
a statistically significantly better accuracy than another algorithm B, 0.5 points if the two algorithms did not have a 
significant difference, and zero points if A was found to be significantly worse than B. 


\textbf{Results.}
In Table~\ref{results-Euclidean} we report the accuracy results. 
We see that SBFIML outperforms in a statistical significant manner the single metric learning method LMNN and the local metric learning methods, GLML and PLML, in seven, eight and six out of ten datasets respectively. When we compare it to KML and SBMML, which learn a Mahalanobis metric in the RKHS and proximity space, respectively, we see that it is significantly better than KML and SBMML in four datasets and significantly worse in one dataset. Compared to $\chi^2$ LMNN, SBFIML outperforms $\chi^2$-LMNN on eight datasets, being statistically significant better on three, and it never loses in statistical significant manner. Finally, compared to SVM, we see that SBFIML is significantly better in two datasets and significantly worse in one dataset. 
In terms of the total score, SBFIML achieves the best predictive performance with 50 point, followed by SVM
,which scores 46.5 point, and $\chi^2$-LMNN with 42 point. The local metric learning method GLML is the one that performs the worst. A potential explanation for the poor performance of GLML could be that its Gaussian distribution assumption is not that appropriate for the datasets we experimented with. 

\begin{table}
\begin{center}
\caption{Accuracy results on large datasets.}
\label{results-large}
\vskip 0.15in
 \scalebox{0.7}{
\begin{tabular}{l||c||c||c}
          
Datasets(\#Inst./\#Feat./\#Class)    & SBFIML    &SBMML    &$\chi^2$ LMNN             \\ \hline \hline
German(1000/20/2)   & 69.40$^{==}$(1.0)& 69.30(1.0)& 69.10(1.0)\\ 
Image(2310/18/2)   & 98.05$^{==}$(1.0)& 98.18(1.0)& 97.79(1.0)\\ 
Splice(3175/60/2)   & \textbf{90.93}$^{+=}$(1.5)& 90.55(0.5)& 90.87(1.0)\\ 
Isolet(7797/617/26)   & 95.45$^{==}$(1.0)& 95.19(1.0)& 95.70(1.0)\\ 
Pendigits(10992/16/10)   & \textbf{98.08}$^{++}$(2.0)& 97.68(0.5)& 97.77(0.5)\\ \hline \hline
Total Score  & 6.5  &  4.0  & 4.5  \\
\end{tabular}
}
\end{center}
\vskip -0.2in
\end{table}

To provide a better understanding of the predictive performance difference between SBFIML, SBMML, and $\chi^2$ LMNN, we applied them on five large datasets. To speedup the learning process, we use as anchor points $20\%$ of randomly selected training instances. Moreover, the parameter $k$ of low rank transformation matrix $\mathbf L$ was reduced to $k=0.05n$, where $n$ is the number of anchor points. The kernel function and similarity map was selected using 4-fold inner CV. The classification accuracy of Isolet and Pendigits are estimated by the default train and test split, for other three datasets we used 10-fold cross-validation. The statistical significance of difference were tested with McNemar's test with p-value of 0.05. 

The accuracy results are reported in Table \ref{results-large}. We see that SBFIML achieves statistical significant better accuracy than SBMML on the two datasets, Splice and Pendigits. When compare it to $\chi^2$ LMNN, we see it is statistical significant better on one dataset, Pendigits. In terms of total score, SBFIML achieves the best score, $6.5$ points, followed by $\chi^2$ LMNN.
\section{Conclusion}
\label{sec:con}

In this paper we present a two-stage metric learning algorithm SBFIML. It first maps learning instances onto a statistical manifold via a similarity-based differential map and then defines the distance in the input data space by the Fisher information distance on the statistical manifold. This induces a new family of distance metrics in the input data space with two important properties. First, the induced metrics are robust to density variations in the original data space and second they have largest distance discrimination on the manifold of the anchor points. Furthermore, by learning a metric on the statistical manifold SBFIML can learn distances on different types of input feature spaces. The similarity-based map used in SBFIML is natural and flexible; unlike KML it does not need to be PSD. In addition SBFIML can be interpreted as a local metric learning method with a well defined distance approximation. The experimental results show that it outperforms in a statistical significant manner both metric learning methods and SVM.


\section*{Acknowledgments}
Jun Wang was partially funded by the Swiss NSF (Grant 200021-122283/1). Ke Sun is partially supported by the Swiss State Secretariat for Education, Research and Innovation (SER grant number C11.0043). Fei Sha is supported by DARPA Award \#D11AP00278  and ARO Award \#W911NF-12-1-0241

\section*{Appendix}
\textbf{Proof of Lemma 2.}
\begin{proof}
Let $\vect \theta$ be the coordinate of $p\in {\mathcal U} \subset \mathcal M^n$ under some smooth coordinate map $\varphi_{\mathcal U}$ and $\vect \gamma$ be the coordinate of $f(p)\in {\mathcal U'} \subset \mathcal N^m$ under some smooth coordinate map $\varphi_{\mathcal U'}$. Since $p' $ approaches $ p $,  we have $\vect \theta'=\vect \theta + d\vect \theta$, where $\vect \theta'$ is the coordinate of $p'$ under the coordinate map $\varphi_{\mathcal U}$ and $d\vect \theta $ is an infinitesimal small change approaching $\vect 0$. Furthermore, since $f:\mathcal M^n \longrightarrow \mathcal N^m$ is a differential map, the function $\varphi_{\mathcal U'}\circ f \circ\varphi^{-1}_{\mathcal U}:\Theta \longrightarrow \Gamma$, which we will denote by $g$, is also differentiable. According to the Taylor expansion, we have $\vect \gamma'=g(\vect \theta')=g(\vect \theta + d\vect \theta) =g(\vect \theta)+ \bigtriangledown g(\vect \theta) d\vect \theta + R_g(d\vect \theta,\vect \theta)=\vect \gamma+ \mathbf J d\vect \theta+ R_g(d\vect \theta,\vect \theta)$, where $\vect \gamma'$ is the coordinate of $f(p')$ under the coordinate map $\varphi_{\mathcal U'}$, $\mathbf J$ is the Jacobian matrix of the function $g$ at point $\vect \theta$ and $R_g(d\vect \theta,\vect \theta)$ is the remainder term of linear approximation. Finally, according to the definition of pullback metric, we have $\lim_{d\vect \theta \to \vect 0} \frac{d_{\mathbf G(\vect \gamma)}(\vect \gamma',\vect \gamma)}{d_{\mathbf G^*(\vect \theta)}(\vect \theta',\vect \theta)}=\lim_{d\vect \theta \to \vect 0}\frac{(\mathbf Jd\vect \theta+R_g(d\vect \theta,\vect \theta))^T\mathbf G(\vect \gamma)(\mathbf Jd\vect \theta+R_g(d\vect \theta,\vect \theta))}{d\vect \theta^T\mathbf J^T\mathbf G(\vect \gamma)\mathbf Jd\vect \theta}=1$. This ends the proof.
\end{proof}

\bibliography{FIML_ICML}
\bibliographystyle{icml2014}

\end{document}